\documentclass[11pt]{article}
\usepackage[final]{acl}

\usepackage{times}
\usepackage{algorithm}    
\usepackage{algorithmic}  
\usepackage{amsmath}      
\usepackage{graphicx}
\usepackage{latexsym}
\usepackage{subcaption}
\usepackage{wrapfig}
\usepackage[normalem]{ulem}
\usepackage[table]{xcolor}
\usepackage[T1]{fontenc}
\usepackage[utf8]{inputenc}
\usepackage{arydshln}
\usepackage{amsfonts}
\usepackage{placeins}
\usepackage{latexsym}
\usepackage{booktabs}
\usepackage{multirow}
\usepackage{inconsolata}
\usepackage{bm}
\usepackage{wrapfig}
\usepackage{arydshln}
\usepackage{mathtools}
\usepackage{amsthm}
\usepackage{url}
\usepackage{hyperref}
\usepackage{enumitem}
\usepackage{xcolor}
\definecolor{lightblue}{RGB}{46,139,87}

\title{Mitigating Position Bias in Transformers via Layer-Specific Positional Embedding Scaling}

\author{
    \textbf{Changze Lv} \thanks{Equal contribution.}\textsuperscript{1,3} \textbf{,} \
    \textbf{Zhenghua Wang} \footnotemark[1]\textsuperscript{1,3}  \textbf{,} \
    \textbf{Yiran Ding} \footnotemark[1]\textsuperscript{2} \textbf{,} \
    \textbf{Yixin Wu} \textsuperscript{1,3}\textbf{,}\
    \textbf{Tianlong Li}\textsuperscript{1,3}\textbf{,}\\
    \textbf{Zhibo Xu} \textsuperscript{1,3}\textbf{,}\
    \textbf{Muling Wu} \textsuperscript{1,3}\textbf{,}\
    \textbf{Tianyuan Shi} \textsuperscript{1,3}\textbf{,}\
    \textbf{Shizheng Li} \textsuperscript{1,3}\textbf{,}\
    \textbf{Qi Qian} \textsuperscript{1,3}\textbf{,} \\
    \textbf{Xuanjing Huang}\textsuperscript{1,3} \textbf{,}\
    \textbf{Xiaoqing Zheng}\thanks{Corresponding Author.} \textsuperscript{1,3}\\
    \textsuperscript{1}Fudan University, \textsuperscript{2}Westlake University \\
    \textsuperscript{3}Shanghai Key Laboratory of Intelligent Information Processing \\
    {\tt\small $\{$zhenghuawang23, czlv24$\}$@m.fudan.edu.cn}
    {\tt\small $\{$yiran.ding$\}$@hdu.edu.cn} \\
    {\tt\small $\{$xjhuang,zhengxq$\}$@fudan.edu.cn}
}

\definecolor{lightpurple}{rgb}{0.9137, 0.9255, 0.9647}
\definecolor{darkpurple}{rgb}{0.8118, 0.8392, 0.9255}
\definecolor{lightgreen}{rgb}{0.8902, 0.9490, 0.8510}
\definecolor{darkgreen}{rgb}{0.7843, 0.8980, 0.7020}
\begin{document}

\maketitle
\begin{abstract}
Large Language Models (LLMs) still struggle with the ``lost-in-the-middle'' problem, where critical information located in the middle of long-context inputs is often underrepresented or lost. 
While existing methods attempt to address this by combining multi-scale rotary position embeddings (RoPE), they typically suffer from high latency or rely on suboptimal hand-crafted scaling strategies. 
To overcome these limitations, we introduce a layer-specific positional embedding scaling~(LPES) method that assigns distinct scaling factors to each layer. 
LPES achieves a more balanced attention distribution without fine-tuning model parameters or increasing inference delay.
A specially designed genetic algorithm is employed to efficiently select the optimal scaling factors for each layer by incorporating B\'{e}zier curves to significantly reduce the search space.
Extensive experiments demonstrate that LPES effectively mitigates positional attention bias and delivers consistent improvements across multiple long-context benchmarks, yielding up to an $11.2$\% accuracy gain on the key-value retrieval dataset.
\end{abstract}

\section{Introduction}

Enabling Large Language Models (LLMs) to process long inputs is essential for supporting complex tasks such as long-text summarization \citep{feng2021survey,zhang2021summ}, code generation \citep{zheng2023codegeex,liu2024your}, and long-context question-answering \citep{li2024long}.
Rotary position embeddings (RoPE) \cite{Su_Lu_Pan_Wen_Liu_2021}, widely adopted in transformer-based LLMs, were designed to encode relative distances between input tokens, facilitating more effective processing of long-context inputs.
However, as the context length increases, RoPE-based LLMs continue to suffer from positional bias. A representative manifestation of this issue is the well-known lost-in-the-middle phenomenon \citep{liu2024lost}, where models tend to over-attend to tokens near the beginning and the end of the input, while relatively neglecting information located in the middle.

Several approaches have been proposed to address the position bias problem by combining multiple RoPEs with different bases or scaling factors \citep{chen2023fortify, zhang2024found, lin2024mixture}.
\citet{chen2023fortify} observed that RoPE with different bases induces attention troughs at specific positions, which impairs the model’s ability to capture the corresponding content.
To mitigate this, they introduced a method, named Attention Buckets, that combines multiple RoPEs with different bases to achieve a more balanced attention distribution. 
Similarly, \citet{lin2024mixture} proposed an MoICE method that assigns multiple RoPE bases to each attention head and aggregates the outputs through a weighted sum.
However, these methods rely heavily on manually designed rules to determine scaling factors or base values, and require multiple forward passes during inference—one for each specific base or scaling factor—followed by ensembling the results. Although some operations can be parallelized, this procedure inevitably increases inference time and computational cost.

\begin{figure*}[t]
\centering
\includegraphics[width=0.95\linewidth]{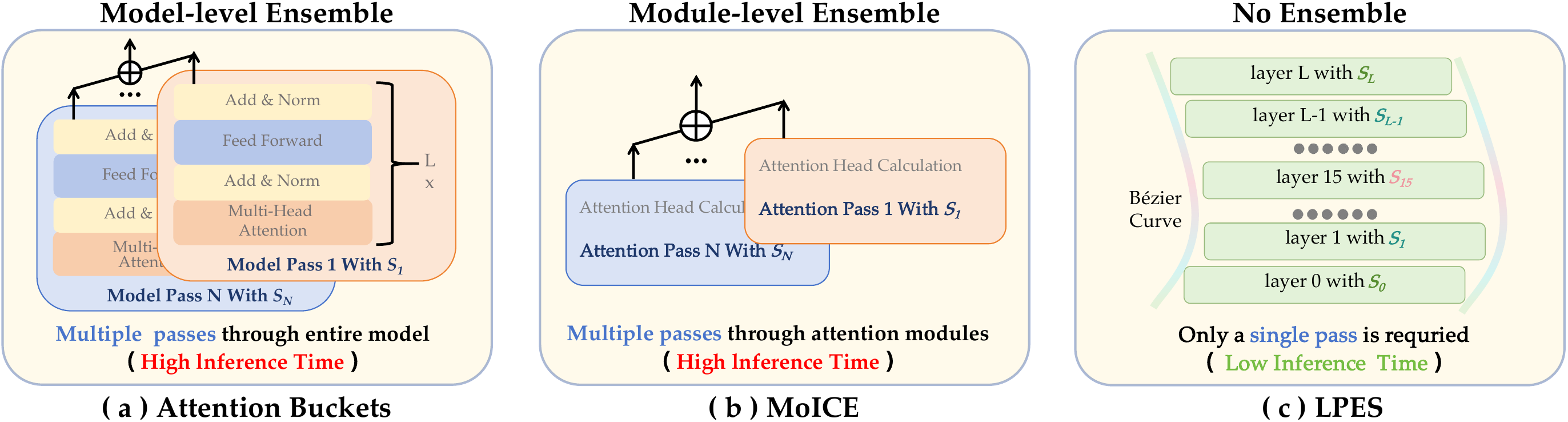}
\caption{
Comparison of the proposed LPES with two representative existing methods. (a) Attention Buckets combines multiple RoPEs with different bases through model parallels. (b) MoICE assigns multiple bases to each attention head.
Unlike these existing methods which require multiple forward passes during inference, our LPES (c) achieves superior performance with a single forward pass, significantly reducing inference time.
}
\label{fig:many_approaches}
\end{figure*}

Varying RoPE bases across the entire model can be seen as model-level ensembling, while applying multiple bases to individual attention heads corresponds to module-level ensembling (Figure~\ref{fig:many_approaches}).
Model-level ensembling requires multiple model inferences, incurring substantial computational overhead, whereas module-level scaling suffers from a large search space due to fine-grained granularity, limiting the applicability of automatic search algorithms. To balance efficiency and flexibility, we apply multiple scaled RoPEs at the layer level, achieving competitive or superior performance with a single forward pass, thus avoiding the associated inference overhead.

Choosing an appropriate scaling factor for each layer is still a non-trivial problem. Let $L$ denote the number of layers in a transformer-based network, and $M$ the number of possible values for the scaling factors; the total number of combinations is $M^L$, which makes an exhaustive search computationally intractable. 
Determining optimal scaling factors is inherently a combinatorial optimization problem, and thus cannot be easily solved by gradient-based methods.
To overcome this, we leverage the B\'{e}zier curve, which defines a smooth, continuous mapping between layer depth and scaling factors using a small set of discrete control points. Letting $C$ denote the number of control points, the search space is reduced to $(M \times L)^C$.
In addition to reducing the search space, we find that the smoothness of curve-based scaling preserves layer-wise representational structure and serves as a beneficial inductive bias.
We further develop a curve-constrained genetic algorithm to solve this combinatorial optimization problem. 
By restricting the search space to B\'{e}zier curves, we can efficiently optimize layer-specific scaling factors, typically within $3$ to $4$ hours using only a few hundred examples (e.g., $200$ instances) on four H100 GPUs.
In long-text tasks, our method introduces no additional inference latency while delivering superior performance over existing approaches.

This study makes the following contributions:
\begin{itemize}[itemsep=2pt, topsep=0pt, leftmargin=*]
\item We propose a layer-specific positional embedding scaling method, termed LPES, which effectively mitigates the position bias without incurring additional inference latency. 
LPES achieves significant speedups, $2.42\times$ faster than MoICE \citep{lin2024mixture} and $1.45\times$ faster than Ms-PoE \citep{zhang2024found}, while also improving the model’s ability to handle long-context tasks. 

\item We introduce an efficient genetic search algorithm in which the search space is constrained by B\'{e}zier curves, enabling rapid optimization of layer-specific scaling factors using only a small set of examples.

\item Extensive experiments on multiple benchmark datasets demonstrate that our method preserves the model’s general capabilities while producing a more balanced attention distribution without costly fine-tuning, making it broadly applicable across different models and tasks.
\end{itemize}

\section{Related Work}

\begin{figure*}[t]
\centering
\includegraphics[width=0.98\linewidth]{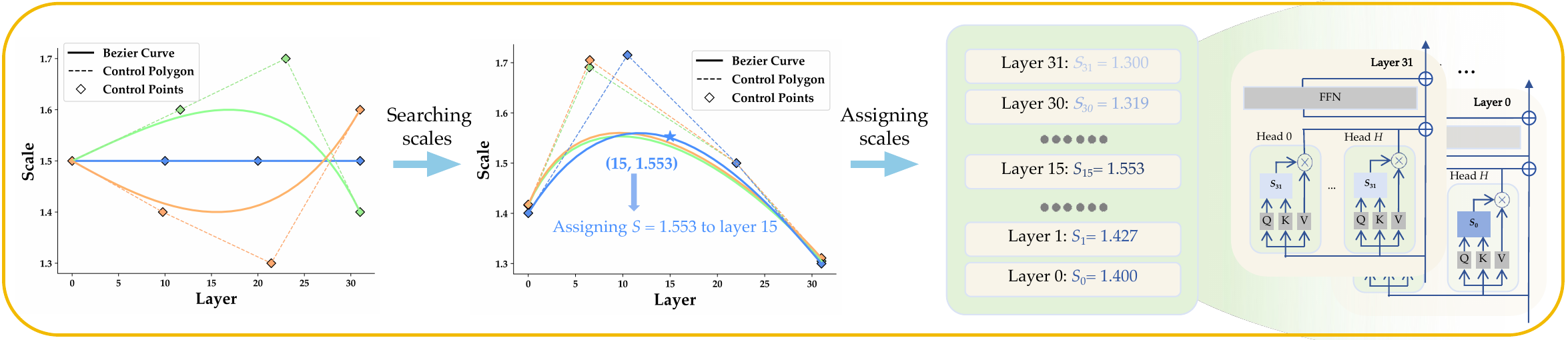}
\caption{\label{fig:work_flow}
Illustration of the proposed layer-specific positional embedding scaling (LPES) method. Left: B\'{e}zier curves can represent a wide variety of shapes. 
Middle: An optimized B\'{e}zier curve found by our search algorithm, which defines a smooth, continuous curve using a limited set of discrete control points.
Right: The relationship between the scaling factors and the optimized B\'{e}zier curve, and their application within the attention mechanism of a transformer-based network. 
}
\end{figure*}

\citet{chen2023fortify} observed that RoPE with different bases can produce attention troughs at specific positions, which is called ``Attention Waves'', thereby impairing the model's ability to capture the relevant content. 
To address this, their ``Attention Buckets'' method integrates multiple RoPE bases through model-parallel inference to achieve a more uniform attention distribution.
\citet{zhang2024found} suggested that the long-term decay in attention may contribute to the position bias, and proposed Ms-PoE that assigns distinct scaling factors to attention heads based on their relative sensitivity to positional information.
MoICE \citep{lin2024mixture}, building on the work of \citet{chen2023fortify}, employs gradient descent to learn the weights for combining results from different bases at the level of individual attention heads.
However, a major limitation of these approaches is their high computational cost and inference latency. Specifically, Attention Buckets requires multiple forward passes, while both Ms-PoE and MoICE require repeated attention computations to integrate multi-scale RoPE information.
They also rely on heuristic or hand-crafted rules to select bases or scaling factors.
By contrast, our method achieves superior performance with a single forward pass and proposes an automatic search algorithm, which effectively determines optimal scaling factors using only a few hundred examples.

\section{Method}

\subsection{Problem Definition} \label{method: Problem Formulation}

In this study, we focus on RoPE, which is defined as follows:
\begin{equation} \label{eq:rope-product-form}
    \langle f({\bm q}, i), f({\bm k}, j) \rangle = {\bm q}^{\mathrm{T}} R(i - j){\bm k}
\end{equation}
\noindent where $f({\bm x}, i)$ denotes a position-dependent rotation applied at position $i$ to the query ${\bm q}$, and $f({\bm k}, j)$ represents the RoPE-rotated key at position $j$.
The notation $ \langle \cdot, \cdot \rangle$ denotes the inner product between the two position-aware vectors, and $R(\Delta)$ is the rotation corresponding to the relative offset $\Delta = i - j$.
This equation shows that the inner product depends only on the vectors $\bm q$, $\bm k$, and the relative distance between them.
\citet{chen2023extending} showed that the context window can be extended by applying a scaling factor $s$ to the position index as follows:
\begin{equation} \label{eq:rope-scaling-form}
    f'({\bm x}, i) = f({\bm x}, i/s) 
\end{equation}
\noindent We further show that the scaling factors can mitigate long-term decay and induce diverse attention patterns (Appendix \ref{appendix:ltd_av}). Accordingly, our goal is to search for a unique scaling factor \(s\) for each layer to combine information from multiple scaled RoPEs, alleviating long-term decay and attention wave effects, and thereby reducing positional bias. 

We model layer depth and scaling factors using Bézier curves, which drastically reduce the search space by determining all layer scales from a few control points. The details are analyzed in Appendix \ref{appendix:search_space}. Furthermore, in Section~\ref{retional_curve}, we show that the smooth and continuous nature of curve-based modeling preserves layer-wise representational structure. Brute-force search demonstrates that smooth scaling naturally emerges as a high-performing configuration, highlighting continuity across layers as a beneficial inductive bias.

As illustrated in Figure~\ref{fig:work_flow}, a B\'{e}zier curve can be viewed as a smooth curve that connects all the scaling factors in a two-dimensional plane.
The problem of selecting scaling factors for all layers can then be transformed into searching for an appropriate B\'{e}zier curve. 
Fortunately, B\'{e}zier curves can model a wide variety of shapes using only a small set of discrete control points, which significantly reduces the search space.
A B\'{e}zier curve of degree $d$, with $d+1$ control points, is defined as follows \citep{mortenson1999mathematics}:
\begin{equation}
    B(t) = \sum_{k=0}^{d} b^d_k(t) P_k, \quad 0 \leq t \leq 1.
\end{equation}
\noindent where $t$ is the parametric coordinate controlling a point's position along the curve, $P_k$ are the control points for the curve, and $b_k^d$ are the Bernstein basis polynomials, which are defined as:
\begin{equation}
    b_k^d(t) = \frac{d!}{k!(d-k)!} t^k (1 - t)^{d-k}, \;\; k = 0, \dots, d.
\end{equation}
\noindent Once a B\'{e}zier curve is determined, the scaling factor $s_h$ for layer $h$ can be computed as follows:
\begin{equation} \label{eq:scaling-factor-setting}
    s_h = \text{proj}_{y} \left[ B( t(x_h) ) \right] 
\end{equation}
\noindent where the notation $\mathrm{proj}_y[\cdot]$ denotes the operation of extracting the $y$-coordinate of a two-dimensional point.
The function $t(\cdot)$ maps $x_h$ to the corresponding parameter $t$ (see Appendix \ref{appendix:t(x)}), where $x_h$ represents the position of layer $h$ within the evenly spaced $x$-coordinates defined by the minimum and maximum values of the control points.
The value of $x_h$ can be computed by:
\begin{equation}
    x_h = P_0^x + \frac{P_d^x - P_0^x}{L-1} \cdot h, \quad h = 0, \dots, L-1.
\end{equation}
\noindent where $L$ denotes the number of layers in a network, and $P_t^x$ is the $x$-coordinates of the $t$-th control point for the B\'{e}zier curve.

Given a training dataset $\mathcal{D} = \{(x_i, y_i)\}_{i=1}^N$ consisting of $N$ examples, where $x_i$ is an input to the large language model and $y_i$ is the corresponding ground-truth output, our goal is to maximize the following function:
\begin{equation} \label{eq:objective-function}
    \mathcal{L}_{\mathcal{D}}({\bm \theta}) = \frac{1}{N} \sum_{i=1}^N \mathbb{I} \{\text{LLM}(x_i, {\bm \theta}) \simeq y_i \}
\end{equation}
\noindent where ${\bm \theta} = (P_0, \dots, P_d)$ denotes the set of control points defining a B\'{e}zier curve of degree $d$ (each control point $P_k$ is a two-dimensional point), $\text{LLM}(x_i, {\bm \theta})$ denotes the output of a language model given input $x_i$, with all scaling factors determined according to Equation (\ref{eq:scaling-factor-setting}) based on the B\'{e}zier curve specified by ${\bm \theta}$, and $\mathbb{I}\{ \cdot \}$ is an indicator function with binary output $0$ or $1$.
We constructed the training dataset such that the content containing information useful for generating correct answers appears at varying positions within the input, thereby encouraging the model to distribute its attention more evenly across the input.

\subsection{Optimization Algorithm} \label{method: Searching the Layer-wise Position Interpolation scales}

We can regard ${\bm \theta} = (P_0, \dots, P_d)$ as a set of newly introduced hyper-parameters that influence the behavior of an LLM.
Each $P_k$ is a two-dimensional vector whose $x$- and $y$-coordinates can take multiple values.
Even though B\'{e}zier curves of degree $d = 3$ which have $d+1 = 4$ control points, are capable of representing a wide variety of curves, Selecting suitable control points is a combinatorial optimization problem that is difficult to solve using gradient-based methods (Appendix~\ref{appendix:gradient}).
Due to the high complexity of the search space, a brute-force approach for determining the scaling factors across layers is intractable; instead, we employ a specialized genetic algorithm to optimize the control points of the B\'{e}zier curves.

In our genetic algorithm, each individual is represented as $(P_0^x, P_0^y, \dots, P_d^x, P_d^y)$, where $P_k^x$ and $P_k^y$ denote the $x$- and $y$-coordinates of the $k$-th control point, and corresponds to a specific B\'{e}zier curve.
The initial population is constructed as follows. First, we initialize an individual in which $k$-th control point is generated by:
\begin{equation} \label{eq:first-individual-generation}
    (k(L - 1)/d, 1.5), \; k \in \{0, \dots, d\}
\end{equation}
\noindent where $L$ is the number of layers in a network. 
Based on the empirical results reported by \citet{zhang2024found}, we set the $y$-coordinate  values of all control points to $1.5$.
Subsequently, the remaining individuals are generated by applying a mutation operator (described below) to this initial individual until the population reaches the predefined size.

The fitness of an individual ${\bm \theta} = (P_0^x, P_0^y, \dots, P_d^x, P_d^y)$ is evaluated by configuring the layer-wise scaling factors of an LLM according to ${\bm \theta}$, running the LLM on a dataset $\mathcal{D}$, and calculating the resulting score $\mathcal{L}_{\mathcal{D}}({\bm \theta})$ as defined in Equation (\ref{eq:objective-function}).
When constructing the training dataset, we deliberately vary the position of relevant context within the input, which can generally be categorized into three types: the query-relevant content appears at the beginning, middle, or end of the input sequence. 
We denote these three corresponding sub-datasets as $\mathcal{D}_\text{B}$, $\mathcal{D}_\text{M}$, and $\mathcal{D}_\text{E}$, respectively.
Considering that original LLMs tend to allocate attention unevenly across different positions, we introduce three weights to reflect the relative importance of these sub-datasets when optimizing the model's scaling factors. 
The final fitness of an individual is then computed as $\lambda_{\text{B}} \mathcal{L}_{\mathcal{D}_\text{B}}({\bm \theta}) + \lambda_{\text{M}} \mathcal{L}_{\mathcal{D}_\text{M}}({\bm \theta}) + \lambda_{\text{E}} \mathcal{L}_{\mathcal{D}_\text{E}}({\bm \theta})$, where $\lambda_{\text{B}} \geq 0$, $\lambda_{\text{M}} \geq 0$, $\lambda_{\text{E}} \geq 0$, and  $\lambda_{\text{B}} +  \lambda_{\text{M}} +  \lambda_{\text{E}} = 1$.

The crossover operator is performed by randomly selecting a pair of individuals with relatively high fitness scores as parents, choosing a single crossover point at random, and exchanging this point between the parents. This process produces two offspring, from which we retain only the one with the higher fitness.

The mutation operator modifies $P^x$ and $P^y$ within a specified range to prevent excessive variations in the resulting curve, as shown in Equation (\ref{constrain2}). 
Let $M_x$ and $M_y$ denote the maximum allowable change for the $x$- and $y$-coordinate, respectively. 
After mutation, the $k$-th control point ($\hat{P}_k^x, \hat{P}_k^y$) of an individual must remain within the following range:

\begin{equation}
\small
\label{constrain2}
\begin{aligned}
\hat{P}_k^x \in
\begin{cases}
\bigl[\max(0, P_k^x - M_x), \min(P_{k+1}^x, P_k^x + M_x)\bigr], 
\\ \qquad \qquad \qquad \qquad \qquad \qquad \qquad \qquad k=0,\\[2pt]
\bigl[\max(P_{k-1}^x, P_k^x - M_x), \min(P_{k+1}^x, P_k^x + M_x)\bigr], 
\\ \qquad \qquad \qquad \qquad \qquad \qquad \qquad \qquad 0<k<d,\\[2pt]
\bigl[\max(P_{k-1}^x, P_k^x - M_x), \min(P_k^x + M_x, L-1)\bigr], 
\\ \qquad \qquad \qquad \qquad \qquad \qquad \qquad \qquad k=d,
\end{cases}
\end{aligned}
\end{equation}

\begin{equation}
\small
\hat{P}_k^y \in \bigl[\max(1, P_k^y - M_y), \min(P_k^y + M_y, 2)\bigr], 0 \le k \le d.
\end{equation}

To ensure the smoothness of the curve and prevent undesirable abrupt changes in the scaling factor \citep{ding2024longRoPE}, the $x$-coordinate values of all control points must increase monotonically. The following condition should therefore be satisfied when performing either crossover or mutation operations. 
Let $P_i^x$ and $P_j^x$ denote the $x$-coordinates of the $i$-th and $j$-th control points, respectively. 
Their relationship is required to satisfy:
\begin{equation} \label{constrain1}
0 \leq P_i^x < P_j^x \leq L-1 \quad \text{if}  \quad i < j 
\end{equation}
\noindent Offspring that fail to meet the above condition are discarded, and the crossover or mutation process is repeated until the condition is satisfied.

Starting with the initial population, individuals are selected based on their fitness, followed by the application of the crossover and mutation operators. This process is repeated iteratively until the maximum number of generations is reached. 
The complete process is summarized in Algorithm \ref{search algorithm}.

\section{Experiments}

The experiments are divided into three parts. First, we evaluate the impact of LPES on context utilization, inference latency, and general capabilities. The results show that LPES improves context utilization while preserving general capabilities and introducing no additional inference latency. Second, we analyze the effectiveness of curve-based modeling for layer-wise scaling factors from the perspectives of inter-layer representational structure and inductive bias. Third, we conduct ablation studies to examine the effects of curve types and the number of control points.

\subsection{Boosting Context Utilization} \label{exp: Enhanced ability to utilize contextual information.}

\begin{table*}[htp]
\centering
\small
\setlength{\tabcolsep}{0.7mm}
\renewcommand\arraystretch{1.0}
\resizebox{1.0\linewidth}{!}{
\begin{tabular}{l|l|ccccc|c|rccccc|c}
\hline
\hline
\bf \multirow{2}{*}{Models} & \bf \multirow{2}{*}{Methods} & $\bf 0\%$ & $\bf 25\%$ & $\bf 50\%$ & $\bf 75\%$ & $\bf 100\%$ & \bf Avg. & $\bf 0\%$ & $\bf 20\%$ & $\bf 40\%$ & $\bf 60\%$ & $\bf 80\%$ & $\bf 100\%$ & \bf Avg. \\
\cline{3-15}
& & \multicolumn{5}{c}{\textbf{MDQA}} & &\multicolumn{6}{c}{\textbf{Key-Value Retrieval}} & \\ \hline

\multirow{6}{*}{Vicuna-7B-v1.5} 
& Baseline 
& $70.4$ & $58.0$ & $55.4$ & $55.4$ & $60.4$ & $59.9$ & $95.2$ & $71.6$ & $81.0$ & $79.0$ & $77.4$ & $73.4$ & $80.9$\\ 
& Positional Interpolation
& $71.2$ & $59.6$ & $58.8$ & $56.4$ & $56.2$ & $60.4$ & $98.6$ & $92.8$ & $83.8$ & $90.0$ & $85.8$ & $83.0$ & $89.0$\\ 
& Attention Buckets
& $\textbf{72.6}$ & $61.4$ & $60.6$ & $60.8$ & $59.6$ & $63.0$ & $\bf 100$ & $\textbf{94.6}$ & $88.6$ & $91.6$ & $87.6$ & $65.8$ & $88.0$\\ 
& Ms-PoE 
& $\textbf{72.6}$ & $61.4$ & $61.8$ & $\textbf{62.0}$ & $59.0$ & $63.5$ & $95.2$ & $63.2$ & $84.8$ & $91.6$ & $87.4$ & $77.8$ & $83.3$\\ 
& MoICE 
& $71.6$ & $61.2$ & $60.6$ & $60.8$ & $\textbf{62.4}$ & $63.3$ & $\bf 100$ & $93.2$ & $\textbf{90.2}$ & $87.4$ & $89.4$ & $70.0$ & $88.4$\\ 
& LPES (Ours) 
& \cellcolor{darkgreen}$71.4$ & \cellcolor{darkgreen}$\textbf{62.2}$ & \cellcolor{darkgreen}$\textbf{62.0}$ & \cellcolor{darkgreen}$61.0$ & \cellcolor{darkgreen}$61.6$ & \cellcolor{darkgreen}$\textbf{63.6}$ & \cellcolor{darkgreen}$99.4$ & \cellcolor{darkgreen}$92.8$ & \cellcolor{darkgreen}$87.8$ & \cellcolor{darkgreen}$\textbf{93.6}$ & \cellcolor{darkgreen}$\textbf{90.4}$ & \cellcolor{darkgreen}$\textbf{88.8}$ & \cellcolor{darkgreen}$\textbf{92.1}$\\ \hline

\multirow{6}{*}{StableBeluga-7B} 
& Baseline 
& $67.8$ & $59.2$ & $59.6$ & $59.4$ & $68.2$ & $62.8$ & $90.2$ & $34.2$ & $44.0$ & $16.6$ & $59.8$ & $79.4$ & $54.0$ \\ 
& Positional Interpolation 
& $\textbf{69.6}$ & $58.6$ & $58.2$ & $60.0$ & $65.4$ & $62.4$ & $95.2$ & $53.6$ & $31.8$ & $28.6$ & $61.6$ & $83.6$ & $59.1$ \\ 
& Attention Buckets
& $69.2$ & $59.0$ & $59.8$ & $59.2$ & $67.4$ & $63.0$ & $\bf 100$ & $79.8$ & $54.4$ & $58.2$ & $68.4$ & $89.2$ & $75.6$ \\ 
& Ms-PoE 
& $68.4$ & $57.0$ & $60.2$ & $\textbf{61.0}$ & $68.4$ & $63.0$ & $90.2$ & $27.2$ & $27.6$ & $\textbf{59.4}$ & $70.4$ & $89.0$ & $60.6$ \\ 
& MoICE 
& $67.4$ & $60.0$ & $60.2$ & $60.0$ & $\textbf{68.6}$ & $63.2$ & $99.8$ & $71.2$ & $52.2$ & $54.8$ & $\textbf{74.4}$ & $91.4$ & $74.0$ \\ 
& LPES  (Ours) 
& \cellcolor{darkgreen}$68.8$ & \cellcolor{darkgreen}$\textbf{60.0}$ & \cellcolor{darkgreen}$\textbf{60.8}$ & \cellcolor{darkgreen}$\textbf{61.0}$ & \cellcolor{darkgreen}$68.2$ & \cellcolor{darkgreen}$\textbf{64.5}$ & \cellcolor{darkgreen}$99.2$ & \cellcolor{darkgreen}$\textbf{82.4}$ & \cellcolor{darkgreen}$\textbf{57.2}$ & \cellcolor{darkgreen}$56.2$ & \cellcolor{darkgreen}$70.4$ & \cellcolor{darkgreen}$\textbf{89.6}$ & \cellcolor{darkgreen}$\textbf{75.8}$ \\ \hline

\multirow{6}{*}{Qwen2.5-7B} 
& Baseline 
& $69.4$ & $61.0$ & $62.6$ & $58.6$ & $63.6$ & $63.0$ & $99.8$ & $88.6$ & $92.6$ & $90.6$ & $99.0$ & $99.2$ & $95.0$ \\ 
& Positional Interpolation  
& $68.6$ & $62.0$ & $62.2$ & $58.4$ & $64.0$ & $63.0$ & $\bf 100$ & $93.2$ & $91.2$ & $88.6$ & $98.6$ & $99.0$ & $95.1$ \\ 
& Attention Buckets
& $69.6$ & $62.2$ & $63.0$ & $60.2$ & $62.0$ & $63.4$ & $\bf 100$ & $89.2$ & $91.4$ & $91.6$ & $98.2$ & $99.2$ & $94.9$ \\ 
& Ms-PoE 
& $69.4$ & $61.8$ & $63.4$ & $60.2$ & $61.4$ & $63.2$ & $\bf 100$ & $94.2$ & $91.2$ & $93.6$ & $98.0$ & $99.2$ & $96.0$ \\ 
& MoICE 
& $68.4$ & $61.2$ & $63.0$ & $61.0$ & $63.8$ & $63.5$ & $99.8$ & $88.0$ & $92.6$ & $91.6$ & $99.0$ & $\textbf{99.4}$ & $95.1$ \\ 
& LPES  (Ours) 
& \cellcolor{darkgreen}$\textbf{69.6}$ & \cellcolor{darkgreen}$\textbf{64.8}$ & \cellcolor{darkgreen}$\textbf{69.2}$ & \cellcolor{darkgreen}$\textbf{63.0}$ & \cellcolor{darkgreen}$\textbf{65.4}$ & \cellcolor{darkgreen}$\textbf{66.4}$ & \cellcolor{darkgreen}$99.8$ & \cellcolor{darkgreen}$\textbf{97.4}$ & \cellcolor{darkgreen}$\textbf{93.2}$ & \cellcolor{darkgreen}$\textbf{94.0}$ & \cellcolor{darkgreen}$\textbf{99.2}$ & \cellcolor{darkgreen}$99.2$ & \cellcolor{darkgreen}$\textbf{97.1}$ \\ 
\hline
\hline
\end{tabular}
}
\caption{Comparison of accuracy across varying positions of relevant information (e.g., $50\%$ denotes the middle) against established baselines. LPES consistently exceeds all baseline performance, validating its efficacy in neutralizing positional bias.}
\label{MDQA_kv}
\end{table*}

\paragraph{Base Models} 
We selected three RoPE-based LLMs for our experiments: Vicuna-$7$B-v$1.5$ \citep{chiang2023vicuna}, and StableBeluga-$7$B \citep{StableBelugaModels}, each with a $4$k-token context window, as well as Qwen$2.5$-$7$B \citep{yang2024qwen2}, which supports a $130$k-token context window.

\paragraph{Benchmarks} 

MDQA \citep{Liu_Lin_Hewitt_Paranjape_Bevilacqua_Petroni_Liang} is a popular multi-document question answering dataset. The key-value retrieval dataset \citep{Liu_Lin_Hewitt_Paranjape_Bevilacqua_Petroni_Liang} features unique UUID key-value pairs, ideal for evaluating relevant information extraction. ZeroSCROLLS \citep{shaham2023zeroscrolls} includes multiple open-ended long-text tasks, with sub-datasets and metrics summarized in Table~\ref{tab:zeroscroll-dataset}. For closed-ended tasks, L-Eval \citep{an2023eval} is used, as outlined in Table~\ref{tab:l-eval-dataset-intro} (Appendix \ref{appendix_dataset}). Finally, MMLU \citep{hendrycks2020measuring} and C-Eval \citep{huang2023c} assess generalization ability across various tasks.

\paragraph{Baselines} 
Positional Interpolation (PI) uses layer-agnostic scaling factors, which are the mean of the searched layer-wise scaling factors \citep{chen2023extending}. 
Attention Buckets performs multiple forward passes, each using a different RoPE base, and then aggregates the information from these passes \citep{chen2023fortify}.
Ms-PoE assigns scaling factors ranging from $1.2$ to $1.8$ to attention heads based on their sensitivity to relevant information \citep{zhang2024found}. 
Building on the work of \citet{chen2023fortify}, MoICE computes attention scores using seven different RoPE bases and then performs a weighted sum of these scores using learned weights \citep{lin2024mixture}.

\paragraph{Experimental Setup} 
For LLMs with a $4$k-token context window, we use $10$ MDQA documents or $50$ key--value pairs as context.
To evaluate positional bias under longer contexts, Qwen$2.5$-$7$B is provided with $20$ MDQA documents or $150$ key--value pairs, and its accuracy is measured as the ground-truth information appears at different positions within the context.
For ZeroSCROLLS and L-Eval, the context window is set to $3{,}584$ tokens, with a maximum of $512$ decoded tokens.
We additionally report the performance of LPES under a $16$K context window in Appendix~\ref{LPES_longer_con}.
To assess generalization, scaling factors learned on MDQA are transferred to ZeroSCROLLS and L-Eval, with additional evaluation on the MMLU and C-Eval benchmarks to measure generalization ability.

During optimization, $\lambda_{\text{B}}$, $\lambda_{\text{M}}$, and $\lambda_{\text{E}}$ are set to $0.2$, $0.3$, and $0.5$, respectively, with detailed analysis provided in Appendix~\ref{Configuration_of_lambda}.
Layer-wise scaling factors are learned by searching the control points of cubic B\'{e}zier curves using $200$ samples from the MDQA or key--value retrieval datasets, and are evaluated on $500$ held-out samples per dataset.

\paragraph{Result Analysis} \textit{Layer-specific positional embedding scaling greatly mitigates position bias.} 
Table~\ref{MDQA_kv} shows that LPES consistently outperforms the baselines in average performance across different positions, notably boosting Vicuna's average accuracy by $11.2\%$ in key-value retrieval.
LPES demonstrates strong transferability when applying MDQA-optimized scaling factors to ZeroSCROLLS and L-Eval (Table~\ref{tab:longtext_horizontal}). The results confirm that these factors generalize robustly across diverse models and long-text tasks.
Furthermore, the results on longer context windows and larger model scales (detailed in Appendix \ref{LPES_longer_con}) further validate the applicability of LPES.
Additionally, LPES preserves the model’s general capabilities, as shown in Table~\ref{tab:general_ability}. Notably, the scaling factors are treated as hyperparameters rather than trainable model parameters. Their adjustment is therefore considered optimization rather than training. In machine learning, training typically refers to updating model weights and biases using gradient-based methods. In contrast, our approach does not modify model parameters and can thus be regarded as a training-free method. This property avoids catastrophic forgetting \cite{de2021continual} caused by large-scale parameter updates and makes the method particularly suitable for already deployed models.

\textit{LPES yields a more balanced attention distribution without additional inference cost.} 
Ms-PoE and MoICE are sample-dependent, their scaling factors cannot be precomputed and must be determined for each input. 
Specifically, Ms-PoE entails an additional attention pass to assess head sensitivity, whereas MoICE requires parallel computations across seven RoPE bases alongside serial routing weight calculations. 

To demonstrate the advantage in inference efficiency, we sample $500$ examples from the MDQA dataset and report the average inference time of Vicuna on a single H100 GPU. 
For a fair comparison, FlashAttention-2~\citep{dao2023flashattention} was used as the attention backend for all methods. 
As shown in Table~\ref{tab:inference_time}, LPES is roughly $1.45\times$ faster than Ms-PoE and $2.42\times$ faster than MoICE.

\begin{table*}[htbp]
    \centering
    \small
    \setlength{\tabcolsep}{0.45mm}
    \renewcommand\arraystretch{1.1}
    \resizebox{1.0\linewidth}{!}{
    \begin{tabular}{l|l|ccccccc|c|cccc|c}
    \hline
    \hline
    & & \multicolumn{8}{c|}{\textbf{Open-ended Long-Text Tasks}} 
      & \multicolumn{5}{c}{\textbf{Closed-ended Long-Text Tasks}} \\
    \cline{3-10} \cline{11-15}
    
    \textbf{Model} & \textbf{Method}
    & \textbf{GovRpt} & \textbf{Qasper} & \textbf{SumScrFd} & \textbf{Qmsum}
    & \textbf{NarrQA} & \textbf{Squality} & \textbf{SpcDgst} & \textbf{Avg.}
    & \textbf{Coursera} & \textbf{QuALITY} & \textbf{TOEFL} & \textbf{SFiction} & \textbf{Avg.} \\
    \hline
    
    \multirow{3}{*}{Vicuna-7B-v1.5}
    & Baseline
    & 18.44 & 22.82 & \textbf{18.42} & 14.50 & 10.98 & 16.56 & 21.39 & 16.91
    & 37.21 & 38.12 & 38.00 & 57.90 & 42.81 \\
    
    & MoICE
    & \textbf{22.29} & 32.34 & 13.31 & 14.79 & \textbf{13.61} & 16.22 & \textbf{22.60} & 19.30
    & \textbf{42.35} & \textbf{43.71} & 39.33 & 57.20 & \textbf{45.65}\\
    
    & LPES (Ours)
    & \cellcolor{darkgreen}21.47 & \cellcolor{darkgreen}\textbf{33.37}
    & \cellcolor{darkgreen}14.39 & \cellcolor{darkgreen}\textbf{15.53}
    & \cellcolor{darkgreen}11.52 & \cellcolor{darkgreen}\textbf{16.91}
    & \cellcolor{darkgreen}22.24 & \cellcolor{darkgreen}\textbf{19.35}
    & \cellcolor{darkgreen}40.41 & \cellcolor{darkgreen}42.57
    & \cellcolor{darkgreen}\textbf{40.67} & \cellcolor{darkgreen}\textbf{58.20}
    & \cellcolor{darkgreen}45.46 \\
    
    \hline
    
    \multirow{3}{*}{Qwen2.5-7B}
    & Baseline
    & 24.76 & 22.92 & 14.69 & 16.25 & 9.78 & 14.85 & 53.66 & 22.42
    & 45.47 & 62.43 & 66.00 & 60.87 & 58.69 \\
    
    & MoICE
    & 25.56 & 23.51 & 15.12 & \textbf{23.19} & 10.64 & \textbf{16.92} & \textbf{53.81} & 24.11
    & 48.13 & 64.28 & 67.33 & 66.00 & 61.44 \\
    
    & LPES (Ours)
    & \cellcolor{darkgreen}\textbf{27.56} & \cellcolor{darkgreen}\textbf{23.91}
    & \cellcolor{darkgreen}\textbf{16.18} & \cellcolor{darkgreen}\textbf{23.19}
    & \cellcolor{darkgreen}\textbf{11.97} & \cellcolor{darkgreen}14.92
    & \cellcolor{darkgreen}\textbf{53.81} & \cellcolor{darkgreen}\textbf{25.51}
    & \cellcolor{darkgreen}\textbf{48.51} & \cellcolor{darkgreen}\textbf{66.43}
    & \cellcolor{darkgreen}\textbf{69.28} & \cellcolor{darkgreen}\textbf{66.42}
    & \cellcolor{darkgreen}\textbf{62.66} \\
    
    \hline
    \hline
    \end{tabular}
    }
    \caption{
    Performance comparison on \textbf{open-ended} and \textbf{closed-ended} long-text benchmarks.
    Open-ended tasks are reported on the left, while closed-ended tasks are shown on the right.
    }

    \label{tab:longtext_horizontal}
\end{table*}

\begin{table}[htp]
    \centering
    \begin{minipage}[h]{1\linewidth}
        \centering
        \small
        \tabcolsep=0.1cm
        \renewcommand\arraystretch{1}
        \begin{tabular}{l|l|cc}
        \hline
        \hline
            \textbf{Model} & \textbf{Method} & \textbf{MMLU} & \textbf{C-Eval} \\ \hline
            \multirow{2}{*}{Vicuna-7B-v1.5} & Baseline & $49.90$  & $49.42$  \\ 
            & LPES (Ours)  & \cellcolor{lightgreen}$49.00$  & \cellcolor{lightgreen}$49.33$ \\ \hline
            \multirow{2}{*}{StableBeluga-7B } & Baseline & $51.50$ & $34.78$   \\
            & LPES (Ours) & \cellcolor{lightgreen}$51.30$  & \cellcolor{lightgreen}$34.63$ \\ 
            \hline
            \hline
        \end{tabular}
        \caption{General capability of models equipped with LPES on MMLU and C-Eval datasets.}
        \label{tab:general_ability}
    \end{minipage}
    \vfill
    \vspace{3mm}
    \begin{minipage}[ht]{1\linewidth}
        \centering
        \small
        \tabcolsep=0.1cm
        \renewcommand\arraystretch{1}
        \begin{tabular}{l|c}
        \hline
        \hline
            \textbf{Method} & \textbf{Inference time per sample (s)} \\ \hline
            Baseline & $0.71$  \\ 
            Attention Buckets & $3.38$  \\
            Ms-PoE & $1.03$  \\
            MoICE & $1.72$  \\ 
            LPES (Ours)  & \cellcolor{lightgreen}$\bf 0.71$ \\ 
        \hline
        \hline
        \end{tabular}
        \caption{Comparison of inference efficiency between LPES and baseline methods.}
        \label{tab:inference_time}
    \end{minipage}
    \vspace{-2mm}
\end{table}

\subsection{Motivation for Curve-Based Modeling}
\label{retional_curve}

\paragraph{Preserved Representational Structure}
The smooth and continuous nature of the B\'{e}zier curve enforces gradual variations in scaling across layers, which helps preserve the coherence of the model’s layer-wise representational structure. We compare against several intuitive baselines: \emph{uniform scaling}, where all layers share a single scaling factor which is the mean of the searched layer-wise scaling factors; \emph{noisy B\'{e}zier scaling}, which adds independent uniform noise sampled from $\mathcal{U}(-0.1, 0.1)$ to each layer’s B\'{e}zier-derived scale; \emph{shuffled scaling}, which randomly permutes the layer-wise scaling factors while preserving their overall distribution; and \emph{fully random scaling}, where each layer independently samples its scale from $\mathcal{U}(1, 2)$.

\begin{figure}[t]
\centering
\includegraphics[width=0.95\linewidth]{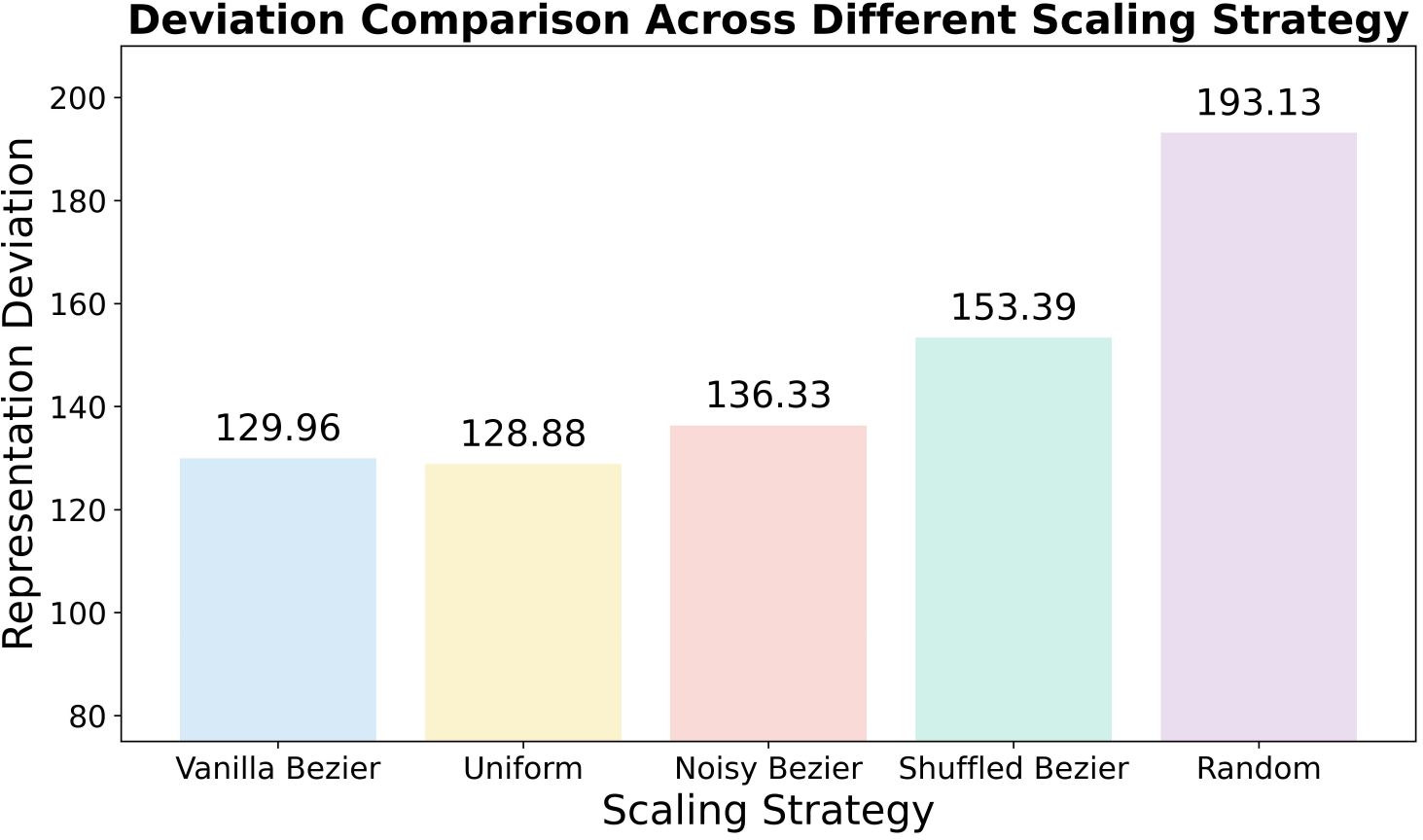}
\caption{
Comparison of representational structure deviation under different scaling strategies.
Vanilla B'{e}zier curve achieves smaller representational deviation while effectively integrating information from multiple base RoPE configurations.
}
\label{fig:dift_of_pe}
\vspace{-2mm}
\end{figure}

Inspired by RSA \cite{kriegeskorte2008representational}, which measures representational structure across input samples at a fixed layer using representational similarity matrices (RSM), we focus instead on the \emph{layer dimension}. 
Specifically, we construct a layer-wise RSM by computing pairwise dot-product similarities between the last-token hidden representations of different layers. 
Given the set of hidden states $\{\mathbf{H}_l\}_{l=1}^{L}$, where $\mathbf{H}_l \in \mathbb{R}^{d}$ denotes the representation at layer $l$, the entries of the RSM are defined as:
\begin{equation}
\mathbf{RSM}_{ij} = \mathbf{H}_i^\top \mathbf{H}_j, \quad 1 \le i, j \le L,
\end{equation}
where $\mathbf{RSM}_{ij}$ quantifies the similarity between the $i$-th and $j$-th layers. This matrix captures the global structural organization of representations across the model's depth.
We quantify representational stability via the \emph{representational structure deviation} $\mathcal{D}$, which measures the average absolute difference between the RSM of a perturbed model ($\mathbf{RSM}^{\text{p}}$) and that of a vanilla configuration without scaling ($\mathbf{RSM}^{\text{v}}$) as follows:
\begin{equation}
\mathcal{D} = \frac{1}{L^2} \sum_{i=1}^{L} \sum_{j=1}^{L} \left| \mathbf{RSM}_{i,j}^{\text{p}} - \mathbf{RSM}_{i,j}^{\text{v}} \right|,
\end{equation}
This metric captures global changes in inter-layer relationships across layers, rather than at individual layers. 
Experiments are conducted using Vicuna-v1.5-7B on 500 randomly sampled MDQA examples, with results averaged over 16 random seeds. 
As shown in Figure~\ref{fig:dift_of_pe}, the smooth B\'{e}zier-based scaling consistently yields smaller structural deviations, indicating minimal disruption to the model’s internal representations.

\begin{figure}[h]
\centering
\includegraphics[width=0.95\linewidth]{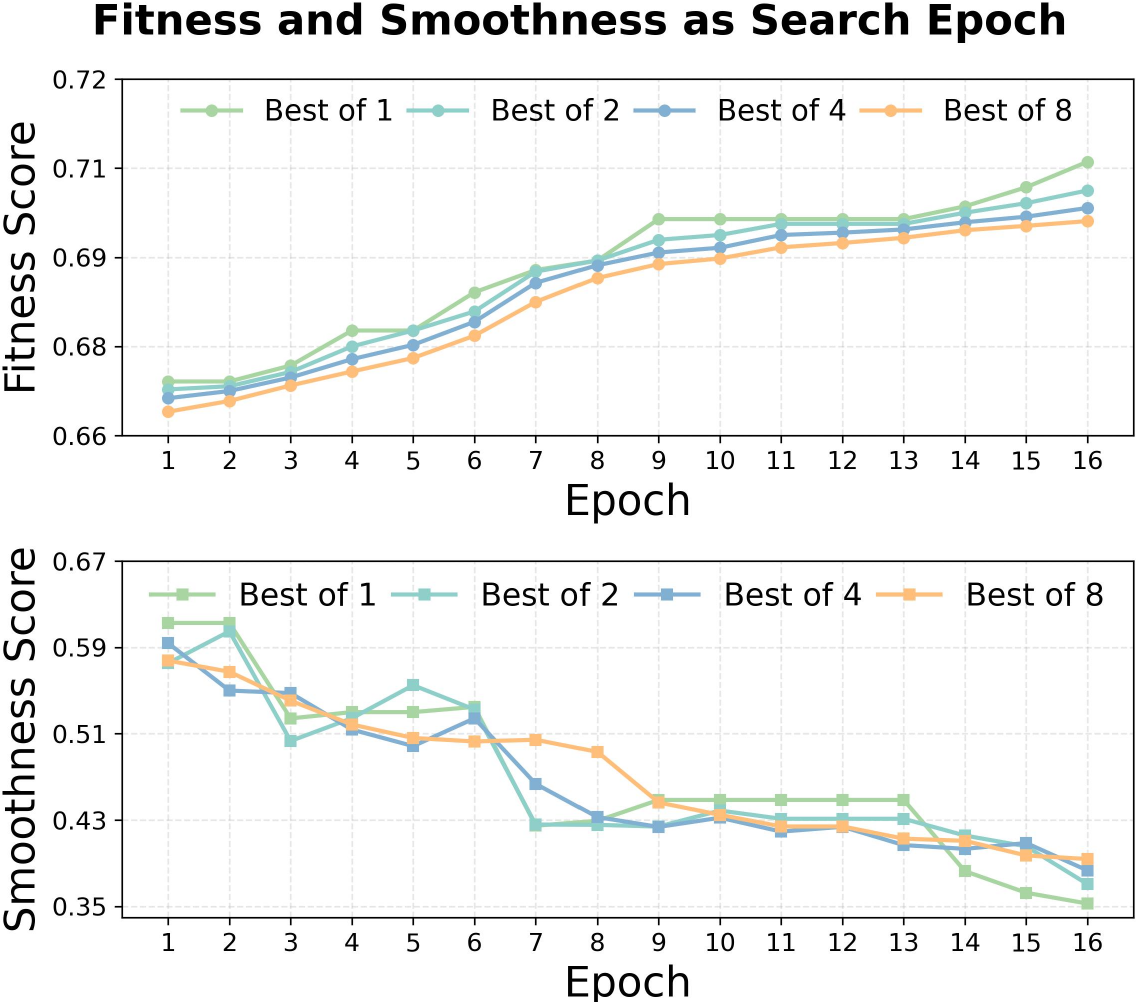}
\caption{
Trend of fitness and smoothness over brute-force search epochs.
The population progressively evolves toward smoother scaling factors across layers, indicating that the smoothness of curves serves as an effective inductive bias.
}
\label{fig:fit_smo_epoch}
\vspace{-2mm}
\end{figure}

\paragraph{Empirical Convergence Behavior} 
We conduct a brute-force genetic algorithm search using the fitness function defined above over $200$ randomly sampled MDQA inputs, where each individual represents a set of layer-wise scaling factors, initialized uniformly from $\mathcal{U}(1,2)$ to avoid introducing any prior smoothness bias. During the search, we track both performance—measured by the average fitness of the best $1$, $2$, $4$, and $8$ individuals—and smoothness, quantified by a second-order metric:
\begin{equation}
\mathcal{S} = \frac{1}{L-2} \sum_{l=2}^{L-1} \left\lVert s_{l+1} - 2s_l + s_{l-1} \right\rVert_2 ,
\end{equation}
where $L$ is the total number of layers, and $s_l$ denotes the scaling factor at layer $l$. Here, $\mathcal{S}$ quantifies the local curvature, with smaller values indicating smoother transitions. 
As shown in Figure~\ref{fig:fit_smo_epoch}, higher-performing configurations consistently exhibit lower $\mathcal{S}$, indicating that smooth variation emerges during the search. This finding suggests that smoothness constitutes a beneficial inductive bias, motivating the use of smooth curve modeling—such as B\'{e}zier curves—to efficiently parameterize high-performing scaling configurations.

\subsection{Ablation Studies}
In this section, we present three ablation studies using Vicuna-v1.5-7B on MDQA. First, we demonstrate that B\'{e}zier curves outperform alternative curves in determining layer-specific scaling factors. Next, we examine the impact of control point counts on convergence quality and speed. 
We further provide an ablation study on the hyperparameter $\lambda$ in Appendix~\ref{appendix:gen_alg}.


\begin{table}[th]
  \centering
  \small
  \setlength{\tabcolsep}{0.5mm}
  \renewcommand\arraystretch{1.0}  
  \begin{tabular}{l|ccccc|c}
    \hline
    \hline
    \textbf{Method} & \textbf{0\%} & \textbf{25\%} & \textbf{50\%} & \textbf{75\%} & \textbf{100\%} & \textbf{Average} \\
    \hline
    Baseline & $70.4$ & $58.0$ & $55.4$ & $55.4$ & $60.4$ & $59.92$ \\
    \hline
    LPES (Linear Curve) & $\textbf{71.8}$ & $61.0$ & $\textbf{62.2}$ & $60.0$ & $60.6$ & $63.12$ \\
    LPES (Step Curve)  & $71.6$ & $60.2$ & $59.4$ & $59.2$ & $60.4$ & $62.16$ \\
    LPES (Bézier curve)   & \cellcolor{darkgreen}$71.4$ & \cellcolor{darkgreen}$\textbf{62.2}$ & \cellcolor{darkgreen}$62.0$ & \cellcolor{darkgreen}$\textbf{61.0}$ & \cellcolor{darkgreen}$\textbf{61.6}$ & \cellcolor{darkgreen}$\textbf{63.64}$ \\
    \hline
    \hline
  \end{tabular}
  \caption{Performance comparison of different curve types for determining layer-wise scaling factors. B\'{e}zier curves achieve superior performance.}
  \label{tab:inference_performance_different_curve}
  \vspace{-3mm}
\end{table}

\paragraph{Curve Type}
\label{Curve_Type}
B\'{e}zier curves provide a compact, low-dimensional parameterization capable of approximating a wide variety of curve shapes \citep{nuntawisuttiwong2021approximation}. 
To demonstrate the advantages of Bézier curve modeling, we consider two alternative approaches with the same number of control points: linear interpolation between control points and step-function modeling based on these control points.
Although these alternatives differ in their curve formulations, they also serve as layer-specific scaling strategies within our framework. While linear interpolation offers slightly higher computational efficiency, which can be neglected in the search procedure (Appendix~\ref{appendix:gen_alg}), we ultimately adopt Bézier curves due to their superior performance.
As shown in Table~\ref{tab:inference_performance_different_curve}, Bézier curves outperform other curve-fitting methods, and the minor additional cost required to determine the scaling factors is fully offset by the inference-time performance gains.

\paragraph{Number of Control Points}
\label{Control_Point_Quantity}

Setting the maximum iterations to $20$, we vary the number of control points to evaluate performance—measured by the mean and variance of accuracy across positions—and convergence speed. While more control points improve B\'{e}zier curve fitting precision and the likelihood of finding optimal scaling factors, they also expand the search space, slowing convergence. 
As shown in Table~\ref{tab:control_points_performance}, using four control points provides a favorable trade-off between performance and convergence speed. In contrast, brute-force search shows little tendency to converge within the limited number of iterations, further demonstrating the efficiency of our curve-constrained genetic algorithm.

\begin{table}[h]
  \centering
  \small
  \setlength{\tabcolsep}{0.5mm}
  \renewcommand{\arraystretch}{1.0} 
  \begin{tabular}{l|c|c}
    \hline
    \hline
    \textbf{Control Points} & \textbf{Accuracy (Std)} & \textbf{Epochs to Convergence} \\
    \hline
    Baseline     & $59.9$ ($\pm 5.56$) & $--$ \\
    \hline
    Brute-Force  & $60.2$ ($\pm 4.69$) & $20$ \\
    \hline
    $2$            & $60.6$ ($\pm 4.55$) & $3$  \\
    $3$            & $62.2$ ($\pm 3.97$) & $5$  \\
    $4$            & $63.6$ ($\pm 3.90$) & $9$  \\
    $5$            & \cellcolor{darkgreen} $\bf 63.8$ ($\bf \pm 3.87$) & \cellcolor{darkgreen}$16$ \\
    \hline
    \hline
  \end{tabular}
  \caption{Performance comparison of different numbers of control points. Additional points improve accuracy and reduce positional bias, but slow convergence due to the increased computational cost of optimization. Overall, experimental results indicate a performance insensitivity to the number of control points used.}
  \label{tab:control_points_performance}
\end{table}

\section{Conclusion}

We present layer-specific positional embedding scaling (LPES), a method that mitigates position bias in transformer-based LLMs by assigning distinct scaling factors to each layer, achieving balanced attention over input without fine-tuning or extra latency. Optimal scaling factors are efficiently identified via a Bézier-constrained genetic algorithm, reducing the search space and converging with only a few hundred examples. Experiments show that LPES consistently improves long-context performance, preserves general capabilities, and requires only a single forward pass, achieving up to $2.42\times$ speedup over MoICE and $1.45\times$ over Ms-PoE, which makes LPES a broadly applicable and efficient solution.

\section*{Limitations}

In this work, we adopt a training-free strategy that assigns different scaling factors across layers to encourage a more balanced attention distribution. While this design enables straightforward and efficient deployment, we do not explore the behavior of our method in training-based settings. Extending the proposed approach to training or fine-tuning pipelines could potentially yield further gains, which we leave for future work. Nevertheless, this limitation does not diminish the practical effectiveness or applicability of our method in real-world scenarios.





\bibliography{custom}
\clearpage

\appendix

\section{Long-Term Decay and Attention Wave in RoPE}
\label{appendix:ltd_av}

\begin{figure*}[htp]
    \centering
    \includegraphics[width=0.9\linewidth]{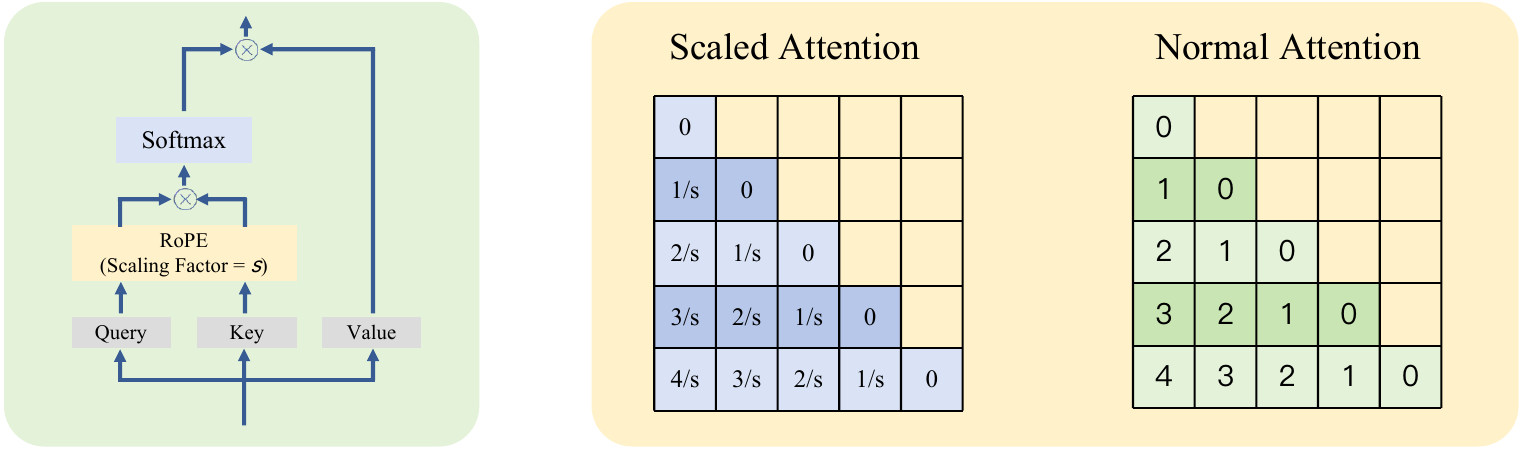}
    \caption{
    We obtain multi-scale RoPE by scaling the positional indices.
    }
    \label{scaled_attention}
\end{figure*}

\begin{figure*}[htp]
    \centering
    \includegraphics[width=0.75\linewidth]{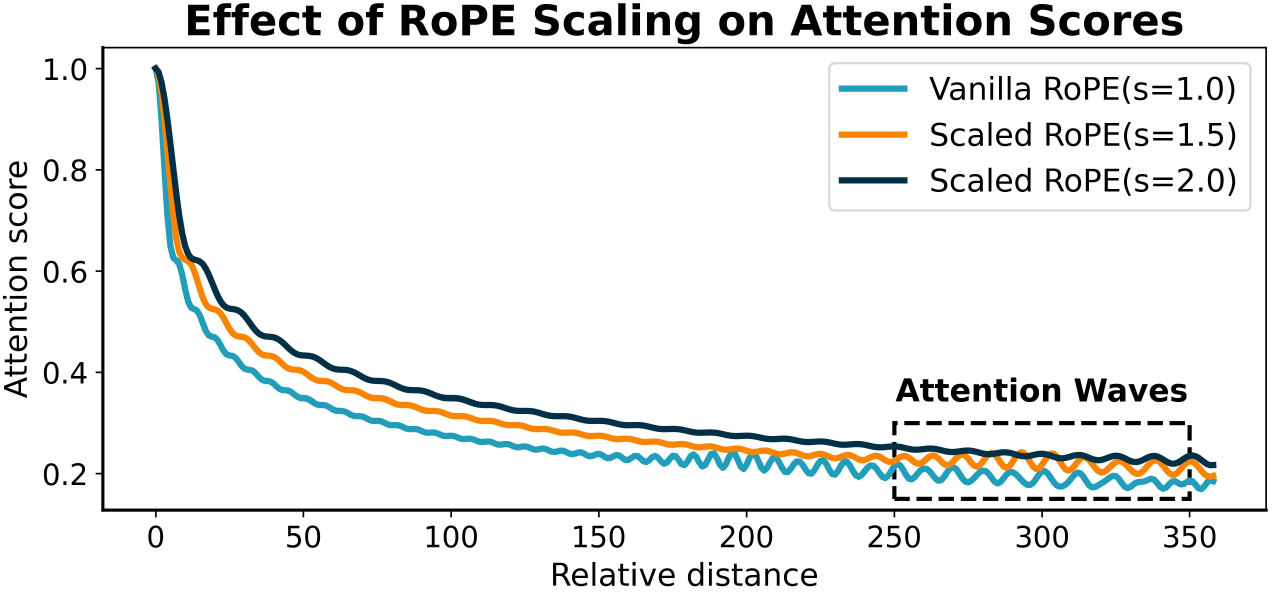}
    \caption{
    The rapid decay of RoPE prioritizes local focus, and the attention waves may cause the model to overlook crucial information at attention troughs, whereas the scaling operation can slow this decay and generate diverse wave patterns.
    }
    \label{long-term-decay}
\end{figure*}

\citet{zhang2024found} observed that the long-term decay of RoPE causes the model to focus more on the end of a sequence. As the relative distance grows, attention scores drop rapidly, leading the model to overemphasize nearby tokens during autoregressive decoding while neglecting distant ones. To mitigate this issue, they scale RoPE by a factor \(s>=1\) (Figure~\ref{scaled_attention}), which effectively reduces the relative distance to \(1/s\) of its original value (Figure~\ref{long-term-decay}). This adjustment slows the decay rate, enabling the model to attend not only to nearby tokens but also to more distant ones, particularly those in the middle of the sequence.

\begin{figure*}[htp]
    \centering
    \includegraphics[width=\linewidth]{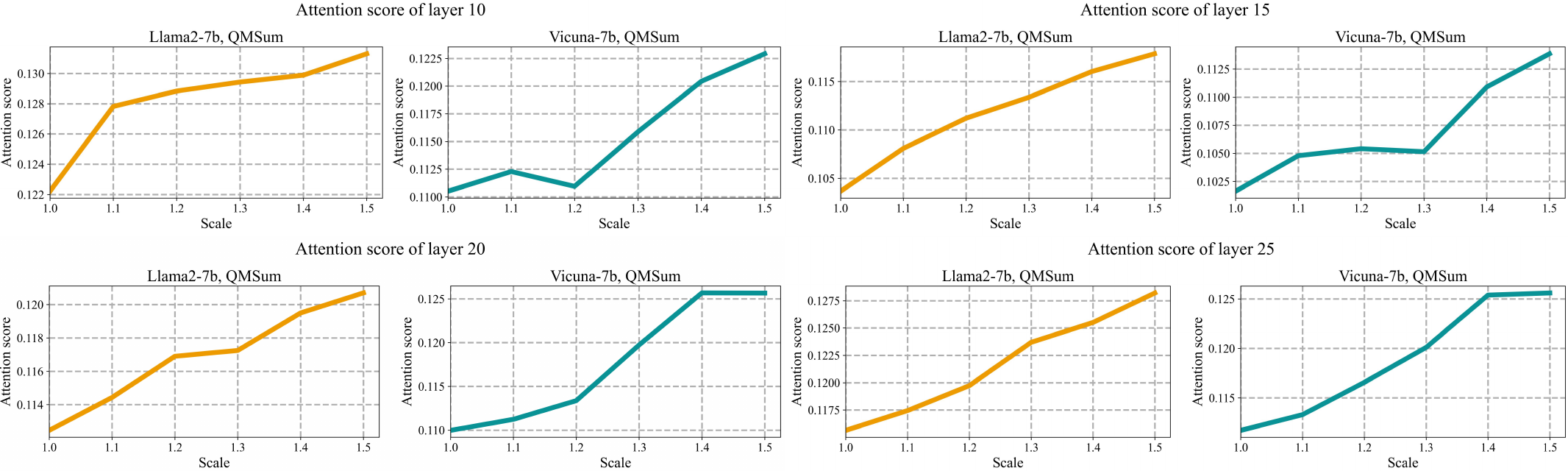}
    \caption{
    The attention score to the middle part across some layers. The scaling operation can enhance the model’s attention to middle positions.
    }
    \label{fig:combined_attention_scores}
\end{figure*}

 To demonstrate that scaling RoPE can indeed enhance the model's attention to middle positions, we use the \textbf{Vicuna-7B-v1.5} \citep{chiang2023vicuna} and \textbf{LLaMA-2-7B-hf} \citep{touvron2023llama} which both consist of 32 transformer layers to conduct experiments on the validation dataset of \textbf{QMSum} \citep{shaham2023zeroscrolls}. We split the context into three parts and calculate the attention scores to the middle-part tokens at different scales.
 In Figure \ref{fig:combined_attention_scores}, an increase in the scale factor leads to higher attention scores, demonstrating that scaling RoPE allows the model to focus more on middle-part content during autoregressive decoding. 

\citet{chen2023fortify} analyze the phenomenon of oscillatory ``attention waves'' in Transformer models, where attention fluctuates across tokens instead of being smoothly distributed. These oscillations, mainly induced by the mechanisms of RoPE, can cause the model to under-attend to important information located at attention troughs, limiting long-context utilization and potentially introducing instability. To address this issue, the authors propose the \textit{Attention Buckets} approach, which runs multiple model parallels with different bases in RoPE and combines the decoded logits across these bases, producing complementary attention wave patterns. The method enhances the model’s sensitivity to context across all positions.

\section{Search Space and Time Complexity Analysis}

\label{appendix:search_space}

We follow \citet{ding2024longRoPE}, discretizing the continuous search space to enable more efficient searching. Assume the control points of the Bézier curve are $(P^x, P^y)$, where $P^x \in [0, L-1]$ ($L$ is the number of scaled layers) and $P^y \in [1, 2] $. The values of $ P^x $ are discretized with a step size of $1$, and the values of $P^y$ are discretized with a step size of $0.1$. Given that the model consists of $32$ layers, there are $32$ possible selections in $P^X$, while the scaling factor chosen from the $P^Y$ set offers $11$ options as shown in Table \ref{search_space}.
 The total number of choices for the brute-force search is $11^{32}$. If a Cubic Bezier curve is used, each control point has $32 \times 11$ possible combinations. With four control points, the total search space is $352^4$ which approximately narrows the search space by a significant factor \textbf{$10^{20}$} compared to the brute-force search.

\begin{table}[h]
\centering
\tiny
\renewcommand\arraystretch{1.1}
\begin{tabular}{c|c}
\hline
\hline
\textbf{Coordinate} & \textbf{Search Space} \\ \hline
\(P^x\) & \{$0$, $1$, $2$, $3$, $4$, $5$, $6$, $7$, $8$,$\dots$, $n-4$, $n-3$, $n-2$, $n-1$\} \\ 
\(P^y\) & \{$1.0$, $1.1$, $1.2$, $1.3$, $1.4$, $1.5$, $1.6$, $1.7$, $1.8$, $1.9$, $2.0$\} \\ \hline \hline
\end{tabular}
\caption{Search space for the control point of Bézier curves.}
\label{search_space}
\end{table}

In our method, the dominant cost of the genetic algorithm arises from evaluating the fitness function, which requires running model inference to assess the effectiveness of different scaling factors. In contrast, the computational overhead of other GA operations—such as assignment, mutation, and crossover—is negligible. Using 4$\times$H100 GPUs, we measured the per-epoch time cost of each operation as follows:

\begin{table}[htp]
  \centering
  \small
  \setlength{\tabcolsep}{4mm}
  \renewcommand\arraystretch{1.1}
  \begin{tabular}{l|c}
    \hline
    \hline
    \textbf{Operation Type} & \textbf{Time (s)} \\
    \hline
    Assigning scaling factor from curve       & $5.2$ \\
    Mutation                                  & $4.5$ \\
    Crossover                                 & $2.3$ \\
    Computing fitness via model inference     & $1167.4$ \\
    \hline
    \hline
  \end{tabular}
  \caption{Measured runtime per epoch of each operation in the genetic algorithm when using 4$\times$H100 GPUs. Model inference dominates the total cost.}
  \label{tab:ga_operation_cost}
\end{table}

Assume the algorithm runs for at most $M$ epochs and generates $N$ new individuals per epoch, and the search uses $S$ samples. Each individual requires three inference runs (placing the correct document at different positions). Thus, the total number of inference calls is $3 N M S$.
In practice, we perform data-parallel inference using $N_{\text{card}}$ GPUs with batch size $B$, which reduces the effective runtime to $O\!\left((3 M N S)/({N_{\text{card}} \cdot B)}\right).$

\begin{table*}[t]
\centering
\setlength{\tabcolsep}{0.1cm} 
\renewcommand{\arraystretch}{1}
\label{tab:ga_hyperparameters}
\begin{tabular}{l c l}
\hline
\hline
\textbf{Hyperparameter} & \textbf{Value} & \textbf{Description}\\
\midrule
\texttt{Population\_size $N_{\text{ps}}$}   & $32$  & Number of individuals in initial population generation.\\
\texttt{Parents\_size $N_{\text{pa}}$}      & $12$  & Number of individuals selected as parents.\\
\texttt{Max\_epoch $T$}         & $20$  & Maximum number of generatios.\\
\texttt{Mutation\_numbers $N_{\text{mu}}$}  & $16$  & Number of offspring generated through mutation. \\
\texttt{Crossover\_numbers $N_{\text{cr}}$} & $4$   & Number of offspring generated through crossover. \\
\texttt{Max\_crossover\_try $N_{\text{ct}}$}& $4$   & Maximum attempts allowed to produce valid offspring during crossover. \\
\texttt{$M_\text{x}$}              & $2$   & Perturbation magnitude of the control point’s $x$-coordinate ($P^x$).\\
\texttt{$M_\text{y}$}              & $0.2$ & Perturbation magnitude of the control point’s $y$-coordinate ($P^y$).\\
\hline
\hline
\end{tabular}
\caption{Hyperparameter settings of the constrained genetic algorithm}
\end{table*}

\section{Limitations of Gradient-Based Methods}
\label{appendix:gradient}

We also attempted to determine the layer-specific scaling factors using gradient descent, but observed poor convergence behavior. This may also shed light on why LongRoPE \citep{ding2024longRoPE} and LongRoPE2 \citep{shang2025longrope2} employ genetic algorithms rather than backpropagation to determine the scaling factors across RoPE dimensions. Although the genetic algorithm incurs higher computational overhead compared to directly optimizing hyperparameters via backpropagation, it consistently converges to a more favorable set of scaling parameters. Furthermore, incorporating Bézier curves significantly accelerates the convergence process.

In the \textbf{gradient-based setting}, we construct three datasets from the MDQA, each containing $2,000$ samples in which the correct document is placed at a different position (i.e., first, middle, or last). In each epoch, a total of $2,000$ samples are drawn from these datasets based on the value of $\lambda$ as specified in Section~\textsection{\ref{exp: Enhanced ability to utilize contextual information.}}, where a larger $\lambda$ indicates a higher probability of sampling from the corresponding dataset. For stable training, we use a batch size of $32$, a learning rate of $1e-5$, and train the model for a total of $30$ epochs.

For the gradient-based method, we observed that even with a large batch size and a small learning rate, the optimization of scaling factors via backpropagation failed to converge. A possible reason is the limited number of trainable parameters \citep{sun2025text}. We evaluated the model at the 30th epoch and found a significant degradation in performance, as shown in Table \ref{gradient_method}. 

\begin{table}[htbp]
\small
\centering
\renewcommand\arraystretch{1.1}  
\setlength{\tabcolsep}{0.5mm} 
{
\begin{tabular}{l|l|ccccc}
\hline
\hline
\bf Model &  \bf Method & \bf 0\% & \bf 25\% & \bf 50\% & \bf 75\% & \bf 100\% \\ \hline
\multirow{2}{*}{Vicuna-7B-v1.5}& Baseline & $70.4$  & $58.0$  & $55.4$  & $55.4$  & $60.4$ \\ 
& Gradient-Based & \cellcolor{darkgreen}$67.4$ & \cellcolor{darkgreen}$54.0$ & \cellcolor{darkgreen}$51.2$ &\cellcolor{darkgreen}$52.8$ &\cellcolor{darkgreen}$55.8$ \\  
\hline
\multirow{2}{*}{Qwen2.5-7B}& Baseline & $69.4$  & $61.0$  & $62.6$  & $58.6$  & $63.6$ \\ 
& Gradient-Based & \cellcolor{darkgreen}$68.7$ & \cellcolor{darkgreen}$56.6$ & \cellcolor{darkgreen}$57.6$ &\cellcolor{darkgreen}$55.8$ &\cellcolor{darkgreen}$57.8$ \\  
\hline
\hline
\end{tabular}
}
\caption{Gradient-based methods lead to accuracy degradation in the MDQA dataset.}
\label{gradient_method}
\end{table}

\section{Cubic Bézier Curve Parameterization for Layer Assignment}
\label{appendix:t(x)}

Consider a cubic Bézier curve with four control points:
\begin{equation}
\begin{aligned}
P_0 &= (x_0, y_0), \quad P_1 = (x_1, y_1), \\
P_2 &= (x_2, y_2), \quad P_3 = (x_3, y_3).
\end{aligned}
\end{equation}
where the $x$-coordinates are strictly increasing since Equation \ref{constrain1}:
\begin{equation}
    x_0 < x_1 < x_2 < x_3.
\end{equation}

The parametric form of the cubic Bézier curve is
\begin{equation} 
    \small
    \label{eq:parametric-form}
    \begin{aligned}
        x(t) &= (1-t)^3 x_0 + 3 (1-t)^2 t x_1 + 3 (1-t) t^2 x_2 + t^3 x_3, \\
        y(t) &= (1-t)^3 y_0 + 3 (1-t)^2 t y_1 + 3 (1-t) t^2 y_2 + t^3 y_3,
    \end{aligned}
\end{equation}
where $t \in [0,1]$.  

Since the $x_i$ are strictly increasing, the function $x(t)$ is typically monotonic. 
This property allows the use of a binary search over the interval $[0,1]$ to efficiently 
find the parameter $t$ corresponding to any given target value $x$, which defines the function $t(x)$.

\section{Hyperparameters of the constrained genetic algorithm}
\label{appendix:gen_alg}

\label{Configuration_of_lambda}
In our experiments, we observed that when scaling RoPE, the model tends to improve performance at early positions while neglecting performance at later positions. Consequently, when setting $\bm \lambda$, we favor assigning larger weights to later positions. Here, we define $\langle \lambda_\text{B}, \lambda_\text{M}, \lambda_\text{E} \rangle$ as the weights assigned to the accuracy of the beginning, middle, and end positions, respectively, in the genetic algorithm's fitness function. In this study, we compare three weighting schemes: $\langle0.333, 0.333, 0.333\rangle$, $\langle0.1, 0.3, 0.6\rangle$, and $\langle0.2, 0.3, 0.5\rangle$.

\begin{table}[hbp]
  \centering
  \small
  \setlength{\tabcolsep}{0.5mm}
  \renewcommand\arraystretch{1.2}  
  \begin{tabular}{l|ccccc|c}
    \hline
    \hline
    \textbf{Method} & \textbf{0\%} & \textbf{25\%} & \textbf{50\%} & \textbf{75\%} & \textbf{100\%} & \textbf{Average} \\
    \hline
    Baseline & $70.4$ & $58.0$ & $55.4$ & $55.4$ & $60.4$ & $59.9$ \\
    \hline
    $\langle 0.333, 0.333, 0.333 \rangle$ & $\textbf{73.2}$ & $\textbf{62.4}$ & $60.2$ & $58.8$ & $58.2$ & $62.6$ \\
    $\langle 0.1, 0.3, 0.6 \rangle$       & $70.6$ & $60.2$ & $60.8$ & $\textbf{61.0}$ & $\textbf{62.0}$ & $63.0$ \\
    $\langle 0.2, 0.3, 0.5 \rangle$       & \cellcolor{darkgreen}$71.4$ & \cellcolor{darkgreen}$62.2$ & \cellcolor{darkgreen}$\textbf{62.0}$ & \cellcolor{darkgreen}$\textbf{61.0}$ & \cellcolor{darkgreen}$61.6$ & \cellcolor{darkgreen}$\textbf{63.2}$ \\
    \hline
    \hline
  \end{tabular}
  \caption{The impact of hyper-parameters $\bm \lambda$ on the optimized layer-wise scaling factors, showing that performance is largely insensitive to their choice.}
  \label{tab:inference_performance_different_weighting}
\end{table}

\section{Search Algorithm Robustness}
In this section, we evaluate the robustness of the scaling factors under variations in the search dataset. On the MDQA dataset, we use Vicuna-1.5-7B and randomly sample $200$ training instances to form the search set for each run. Across five independent runs with different search sets, the method achieves an average performance of $63.68$ with a sample variance of only $0.027$, demonstrating that our approach is highly stable across different search sets. Overall, our method consistently outperforms prior approaches, highlighting the robustness of the proposed search algorithm.

\begin{table}[h]
  \centering
  \small
  \setlength{\tabcolsep}{0.5mm}
  \renewcommand\arraystretch{1.0}  
  \begin{tabular}{l|ccccc|c}
    \hline
    \hline
    \textbf{Method} & \textbf{0\%} & \textbf{25\%} & \textbf{50\%} & \textbf{75\%} & \textbf{100\%} & \textbf{Average} \\
    \hline
    Baseline & $70.4$ & $58.0$ & $55.4$ & $55.4$ & $60.4$ & $59.9$ \\
    Attention Buckets & $\textbf{72.6}$ & $61.4$ & $60.6$ & $60.8$ & $59.6$ & $63.0$ \\
    Ms-PoE & $\textbf{72.6}$ & $61.4$ & $61.8$ & $62.0$ & $59.0$ & $63.5$ \\
    MoICE & $71.6$ & $61.2$ & $60.6$ & $60.8$ & $\textbf{62.4}$ & $63.3$ \\
    \hline
    LPES (run 1) & \cellcolor{darkgreen}$71.4$ & \cellcolor{darkgreen}$62.2$ & \cellcolor{darkgreen}$62.0$ & \cellcolor{darkgreen}$61.0$ & \cellcolor{darkgreen}$61.6$ & \cellcolor{darkgreen}$63.6$ \\
    LPES (run 2) & \cellcolor{darkgreen}$71.6$ & \cellcolor{darkgreen}$62.4$ & \cellcolor{darkgreen}$\textbf{62.2}$ & \cellcolor{darkgreen}$60.8$ & \cellcolor{darkgreen}$61.8$ & \cellcolor{darkgreen}$63.8$ \\
    LPES (run 3) & \cellcolor{darkgreen}$71.6$ & \cellcolor{darkgreen}$61.8$ & \cellcolor{darkgreen}$61.8$ & \cellcolor{darkgreen}$62.0$ & \cellcolor{darkgreen}$61.0$ & \cellcolor{darkgreen}$63.6$ \\
    LPES (run 4) & \cellcolor{darkgreen}$\textbf{72.6}$ & \cellcolor{darkgreen}$61.0$ & \cellcolor{darkgreen}$62.0$ & \cellcolor{darkgreen}$\textbf{63.2}$ & \cellcolor{darkgreen}$61.0$ & \cellcolor{darkgreen}$\textbf{63.9}$ \\
    LPES (run 5) & \cellcolor{darkgreen}$72.2$ & \cellcolor{darkgreen}$\textbf{62.8}$ & \cellcolor{darkgreen}$61.0$ & \cellcolor{darkgreen}$61.0$ & \cellcolor{darkgreen}$60.4$ & \cellcolor{darkgreen}$63.5$ \\
    \hline
    \hline
  \end{tabular}
  \caption{Performance of LPES across five runs compared with baseline methods. Percentages indicate the relative position of relevant documents in the context.}
  \label{tab:lpes_runs}
\end{table}

\section{Dataset Details}
\label{appendix_dataset}

\begin{figure}[htbp]
    \centering
    \begin{subfigure}[htbp]{0.98\linewidth}
        \centering
        \includegraphics[width=\linewidth]{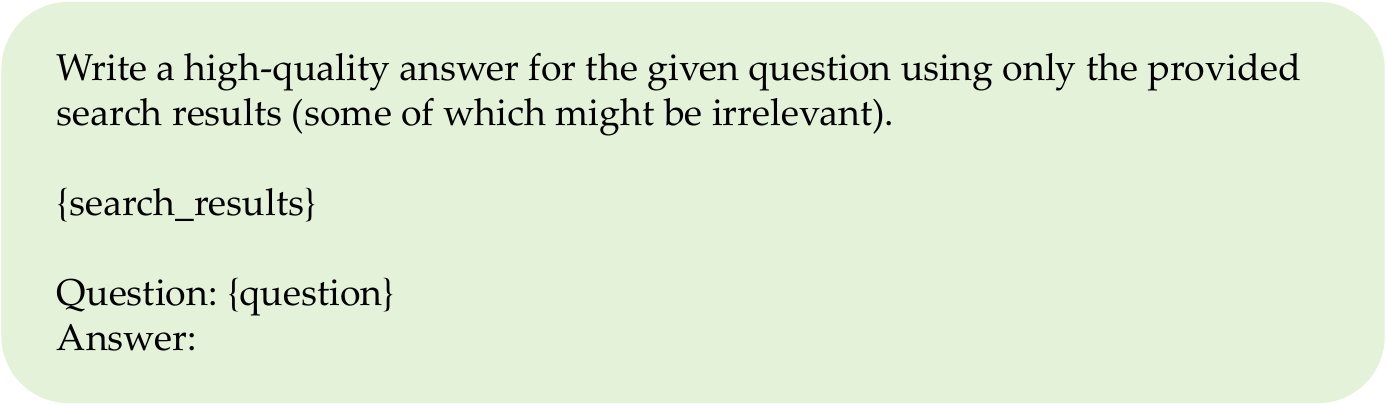}
        \caption{MDQA prompt}
        \label{fig:MDQA_prompt}
    \end{subfigure}
    \vspace{5mm}
    \begin{subfigure}[htbp]{0.98\linewidth}
        \centering
        \includegraphics[width=\linewidth]{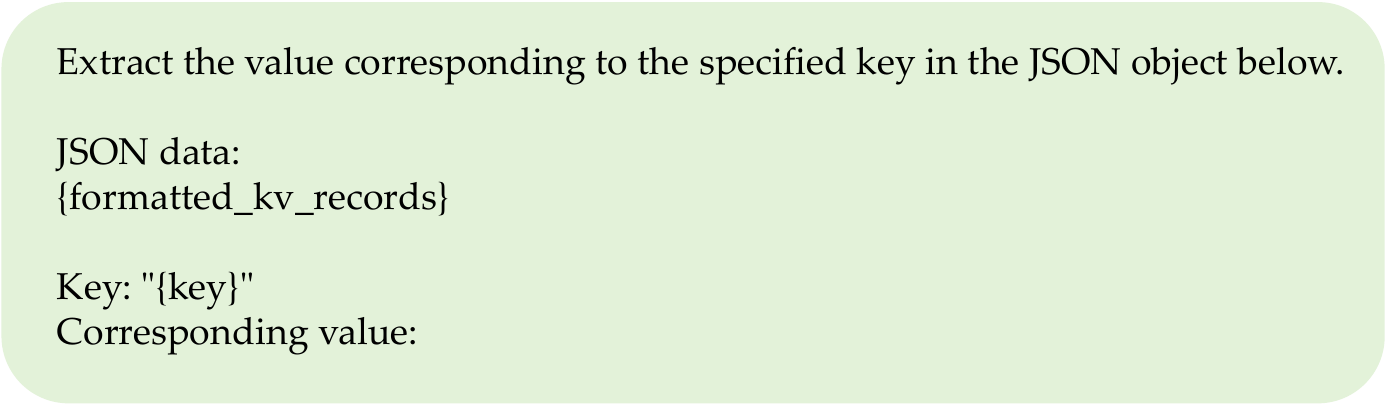}
        \caption{KV prompt}
        \label{fig:KV_prompt}
    \end{subfigure}
    \caption{Prompt templates used in MDQA and Key-Value Retrieval datasets.}
    \label{fig:combined_prompts}
\end{figure}

\begin{table}[h]
    \centering
    \small
    \setlength{\tabcolsep}{0.5mm} 
    \renewcommand{\arraystretch}{1.2} 
    \begin{tabular}{l|l|l|l}
        \hline
        \hline
        \textbf{Dataset} & \textbf{Question Style} & \textbf{Domain} & \textbf{Metric} \\ \hline
        Coursera   & Multiple Choice        & Advanced Courses   & Accuracy \\
        \hline
        QuALITY    & Multiple Choice        & Gutenberg          & Accuracy \\
        \hline
        TOEFL      & Multiple Choice        & English Test       & Accuracy \\
        \hline
        SFiction   & True/False Questions   & Scientific Fiction & Accuracy \\
        \hline
        \hline
    \end{tabular}
    \caption{Overview and evaluation metrics of the sub-datasets in L-Eval.}
    \label{tab:l-eval-dataset-intro}
\end{table}

\begin{table}[h]
    \small
    \centering
    \setlength{\tabcolsep}{0.2mm} 
    \renewcommand{\arraystretch}{1.2}
    \begin{tabular}{l | p{3.5cm} | l}
        \hline
        \hline
        \textbf{Dataset} & \textbf{Description} & \textbf{Metric} \\ 
        \hline
        GovReport     & Summarization of long reports & ROUGE-1/2/L \\ 
        \hline
        SummScreenFD  & Summarization of TV show episode scripts & ROUGE-1/2/L \\
        \hline
        QMSum         & Query-based summarization over meeting transcripts & ROUGE-1/2/L \\
        \hline
        SQuALITY      & Question-focused summarization over stories & ROUGE-1/2/L \\
        \hline
        Qasper        & Question answering over research papers & F1 \\
        \hline
        NarrativeQA   & Question answering about entire books and movie scripts & F1 \\
        \hline
        SpaceDigest   & Aggregated sentiment classification over 50 hotel reviews from Space & Exp\_similarity \\ 
        \hline
        \hline
    \end{tabular}
    \caption{Overview and evaluation metrics of the sub-datasets in ZeroSCROLLS.}
    \label{tab:zeroscroll-dataset}
    
\end{table}


\begin{table*}[htbp]
    \centering
    \small
    \setlength{\tabcolsep}{2mm}
    \renewcommand\arraystretch{1}
    \begin{tabular}{l|l|cccc|c}
    \hline
    \hline
        \textbf{Model} & \textbf{Method} & \textbf{Coursera} & \textbf{QuALITY} & \textbf{TOEFL} & \textbf{SFiction} & \textbf{Average} \\ 
        \hline
        \multirow{3}{*}{Vicuna-13B-v1.5-16k} 
        & Baseline & $69.6$ & $51.4$ & $33.3$ & $57.1$ & $52.9$ \\
        & MoICE & $67.4$ & $\textbf{55.6}$ & $35.7$ & $52.6$ & $52.8$ \\
        & LPES (Ours) 
            & \cellcolor{darkgreen}$\textbf{70.6}$
            & \cellcolor{darkgreen}$54.4$
            & \cellcolor{darkgreen}$\textbf{36.0}$
            & \cellcolor{darkgreen}$\textbf{59.4}$
            & \cellcolor{darkgreen}$\textbf{55.1}$ \\
        \hline
        
        \multirow{3}{*}{Qwen2.5-7B} 
        & Baseline & $59.8$ & $66.3$ & $76.6$ & $71.8$ & $68.6$ \\
        & MoICE & $59.8$ & $66.3$ & $78.7$ & $\textbf{73.0}$ & $69.5$ \\
        & LPES (Ours)
            & \cellcolor{darkgreen}$\textbf{63.8}$
            & \cellcolor{darkgreen}$\textbf{69.1}$
            & \cellcolor{darkgreen}$\textbf{80.9}$
            & \cellcolor{darkgreen}$72.9$
            & \cellcolor{darkgreen}$\textbf{71.7}$ \\
        
    \hline
    \hline
    \end{tabular}
    \caption{Results under longer-context settings (16k tokens) on the L-Eval benchmark. LPES consistently improves performance over the baseline and MoICE on both Vicuna-13B-v1.5-16k and Qwen2.5-7B, demonstrating strong scalability to larger models and longer context windows.}
    \label{tab:long_context_results}
\end{table*}

\begin{table*}[htbp]
    \centering
    \small
    \setlength{\tabcolsep}{2mm}
    \renewcommand\arraystretch{1}
    \begin{tabular}{l|l|ccccccc|c}
    \hline
    \hline
        \textbf{Model} & \textbf{Method} 
        & \textbf{GovRpt} 
        & \textbf{Qasper} 
        & \textbf{SumScrFd} 
        & \textbf{QMSum} 
        & \textbf{NarrQA} 
        & \textbf{SQuality} 
        & \textbf{SpcDgst} 
        & \textbf{Average} \\ 
        \hline
        \multirow{3}{*}{Qwen2.5-7B} 
        & Baseline 
            & $25.17$ & $23.33$ & $14.23$ & $17.36$ & $6.00$ & $17.60$ & $51.12$ & $22.12$ \\
        & MoICE 
            & $\textbf{38.77}$ & $18.83$ & $15.21$ & $17.84$ & $6.14$ & $18.06$ & $53.63$ & $24.07$ \\
        & LPES (Ours) 
            & \cellcolor{darkgreen}$35.77$
            & \cellcolor{darkgreen}$\textbf{24.63}$
            & \cellcolor{darkgreen}$\textbf{15.91}$
            & \cellcolor{darkgreen}$\textbf{20.84}$
            & \cellcolor{darkgreen}$\textbf{8.02}$
            & \cellcolor{darkgreen}$\textbf{19.06}$
            & \cellcolor{darkgreen}$\textbf{53.77}$
            & \cellcolor{darkgreen}$\textbf{25.43}$ \\
        \hline
        \hline
    \end{tabular}
    \caption{Performance comparison on ZeroSCROLLS benchmarks with a 16k context length using Qwen2.5-7B. LPES consistently improves average performance over both the baseline and MoICE across diverse tasks.}
    \label{tab:qwen_long_context}
\end{table*}

\section{Effectiveness of LPES on Longer Contexts}
\label{LPES_longer_con}

We conduct experiments on Vicuna-1.5-13B and Qwen-2.5-7B under a 16k-token context setting on L-Eval to verify the effectiveness of LPES in long-context scenarios. The decoding length is set to $512$ tokens, so the maximum usable context window is limited to $15{,}872$ tokens. As shown in Table~\ref{tab:long_context_results} and \ref{tab:qwen_long_context}, the results demonstrate that our method remains effective on larger models and extended context lengths, highlighting its strong scalability and robustness.

\begin{algorithm*}[htbp]
	\small
	\caption{Scaling factor search algorithm}
    \label{search algorithm}
	\textbf{Input:} an LLM $\mathcal{M}$, a dataset $\mathcal{D}$, population size $N_{\text{ps}}$, the number of offspring generated by crossover $N_\text{cr}$, \\
    {\color{white} zzzzzz} the number of mutated individuals $N_\text{mu}$, and maximum number of generations $T$. \\
	\vspace{-3ex}
\label{appendix:alg}
	\begin{algorithmic}[1]
		\STATE $\mathcal{S}_0$ = \text{Initial-Population-Generation}($\mathcal{D}$, $N_{\text{ps}}$); {\color{lightblue} // Randomly generate the initial population.}
		\FOR{$i = 1$ to $T$}
            \STATE \text{Evaluate-Fitness}($\mathcal{S}_{i-1}$, $\mathcal{M}$, $\mathcal{D}$); {\color{lightblue} // Evaluate the fitness of all individuals in the population.}   
		\STATE  $\mathcal{S}_{\text{pa}}$ = \text{Select-Parents}($\mathcal{S}_{i-1}$); {\color{lightblue} // Select the parent pool according to fitness values.} 
		\STATE $\mathcal{S}_{\text{cr}}$ = \text{Crossover-Operator}($\mathcal{S}_{\text{pa}}$, $N_{\text{cr}}$); {\color{lightblue} // Produce offspring using the crossover operator.} 
        \STATE$\mathcal{S}_{\text{mu}}$ = \text{Mutation-Operator}($\mathcal{S}_{\text{pa}}$, $N_{\text{mu}}$); {\color{lightblue} // Generate offspring using the mutation operator.} 
		
		\STATE $\mathcal{S}_i$ = $\mathcal{S}_{\text{pa}}$ $\cup$ $\mathcal{S}_{\text{cr}}$ $\cup$ $\mathcal{S}_{\text{mu}}$; {\color{lightblue} // Merge the individuals to form the next generation's population.} 
		\ENDFOR
		\STATE Return the individual with the highest fitness in $\mathcal{S}_{T}$.
	\end{algorithmic}  
    
\end{algorithm*}


\end{document}